\documentclass[nonacm,sigconf]{acmart}

\AtBeginDocument{%
  \providecommand\BibTeX{{%
    \normalfont B\kern-0.5em{\scshape i\kern-0.25em b}\kern-0.8em\TeX}}}

\setcopyright{acmcopyright}
\copyrightyear{2020}
\acmYear{2020}
\acmDOI{10.1145/1122445.1122456}

\acmConference[Workshop on NLG for HRI]{Workshop on Natural Language Generation for Human-Robot Interaction}{March 23, 2020}{Cambridge, UK}
\acmBooktitle{Workshop on NLG for HRI: Workshop on Natural Language Generation for Human-Robot Interaction,
  March 23, 2020, Cambridge, UK}
\acmPrice{15.00}
\acmISBN{978-1-4503-XXXX-X/18/06}

\usepackage{booktabs}
\usepackage{multirow}
\usepackage{algpseudocode}
\usepackage{float}
\usepackage{subfigure}
\usepackage[linesnumbered,ruled]{algorithm2e}
\usepackage{graphicx}
\usepackage{textcomp}
\usepackage{xcolor}
\usepackage{stackengine}
\usepackage{pgfplots}
\usepackage{pgfplotstable}
\usepackage{ifthen}
\pgfplotsset{compat=1.7}
\usepgfplotslibrary{groupplots}

\begin{document}

\title{Open Challenges on Generating Referring Expressions for Human-Robot Interaction}

\author{Fethiye Irmak Do\u{g}an}
\affiliation{%
  \institution{KTH Royal Institute of Technology}
  \city{Stockholm}
  \country{Sweden}
}
\email{fidogan@kth.se}

\author{Iolanda Leite}
\affiliation{%
  \institution{KTH Royal Institute of Technology}
  \city{Stockholm}
  \country{Sweden}
}
\email{iolanda@kth.se}
\renewcommand{\shortauthors}{F. I. Do\u{g}an and I. Leite}

\begin{abstract}
Effective verbal communication is crucial in human-robot collaboration. When a robot helps its human partner to complete a task with verbal instructions, referring expressions are commonly employed during the interaction. Despite many studies on generating referring expressions, crucial open challenges still remain for effective interaction. In this work, we discuss some of these challenges (i.e., using contextual information, taking users' perspectives, and handling misinterpretations in an autonomous manner).
\end{abstract}





\maketitle
\pagestyle{empty}
\section{Introduction}
\begin{figure}
    \centering
    \includegraphics[width=0.49\textwidth]{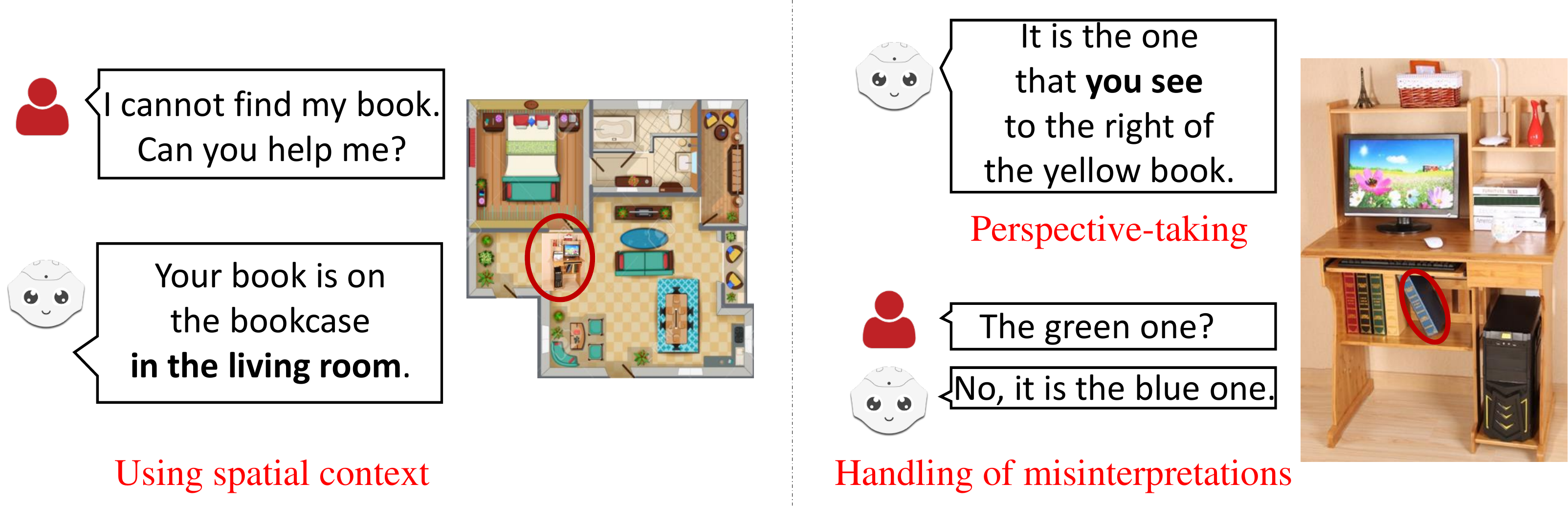}
    \caption{Example interactions for open challenges on generating referring expressions.}
    \label{fig:overview}
\end{figure}
In a human-robot collaborative task, it is critical that verbal communication between a robot and a human is effective. For instance,  when a human assembles furniture, and a robot helps to find the correct pieces, the robot should direct its human partner and describe the target objects effectively. Expressions used for describing objects in terms of their distinguishing features are called referring expressions, and referring expression generation is defined as \textit{``choosing the words and phrases to express domain objects''} \cite{foster2019natural}.

Generating appropriate referring expressions has the potential of significantly improving human-robot collaboration. It is one of the most studied areas in natural language generation for social robotics because the problem contains a relatively straightforward input and output \cite{foster2019natural}. Studies on referring expression generation based on primarily rule-based templates or algorithms \cite{williams2017referring,williamsreferring,kunze2017spatial,zender2009situated}, and recent studies have addressed this problem using learning-based methods \cite{dougan2019learning,magassouba2019multimodal,tanaka2019generating}.

Although referring expression generation has been extensively studied in HRI, there are still open challenges for more efficient generation mechanisms applicable to different tasks. In this paper, after reviewing the state of the art research in this area, we summarize the main open challenges and open further research directions (see Figure \ref{fig:overview})
, which can be summarized as follows:
\begin{itemize}
    \item Leveraging contextual information to generate referring expressions that facilitate communication 
    \item While generating referring expressions, taking the perspectives of users for an effective collaboration
    \item Being able to complete the task accurately in dynamic environments by autonomously handling the misinterpretations
\end{itemize}
\section{Open Challenges}
\subsection{Using Contextual Information}
When a robot describes an object to its human partner, it needs to consider the social context (e.g., the person's age and knowledge level about the object) and spatial context (e.g., whether it is likely to find the target object in the existing place or the robot needs to direct the user to another place), and adapt itself accordingly. 

Understanding social context is essential to generate comprehensible referring expressions. For instance, if a robot describes a rarely known object to a child, it can be more efficient to use color or shape information of the object instead of solely using the object's name.

To facilitate finding the described object, interpreting spatial context is crucial. For example, if a robot describes an object which is more likely to belong to a kitchen,
it can direct the user to the correct place with its referring expressions, e.g., "the object A next to the object B \textbf{in the kitchen}".
Otherwise, users can waste time by looking for objects in the wrong places, which can affect the user's perception of the interaction and the robot. To address these challenges, recent studies on context modeling in robots have employed deep learning \cite{dougan2018cinet,dougan2018deep,8460828,bozcan2019cosmo}. In these studies, co-occurrences of the objects \cite{dougan2018deep}, spatial relations between them \cite{8460828}, and their affordances \cite{bozcan2019cosmo} have been utilized.

Adapting referring expressions concerning context has been focused \cite{viethen2010speaker,garoufi2014generation,krahmer2002efficient,foster2014task}. In a promising referring expression study, Viethen and Dale \cite{viethen2010speaker} showed that human behavior on selecting the content of referring expressions are mostly speaker-dependent. They stated that this speaker-dependency might be correlated to ``age, gender, and social or cultural background'' of the users, and this is open for further research. Some studies have focused on context-dependent selection of distinguishing features of objects \cite{viethen2010speaker,garoufi2014generation,krahmer2002efficient} rather than using fixed preference ordering over features \cite{dale1995computational}. Further, Foster et al. \cite{foster2014task} studied the context-sensitive generation of referring expressions on human-robot joint construction tasks. Their context-sensitive expressions include "... this red cube and screw it ..." where "this red cube" and "it" are called context-sensitive references. Even though these works have important improvements in context-dependency of referring expressions, considering all aspects of social and spatial contexts remains unsolved.


\subsection{Perspective-Taking}

Perspective-taking has been a research topic in a variety of fields (e.g., psychology, cognitive science, robotics, and computer vision). In robotics, it has been studied when the robot comprehends referring expressions \cite{berlin2006perspective,wiltshire2013towards,pandey2010mightability}. Many of these studies focused on perspective-taking when users employ spatial relations \cite{fong2005peer,kennedy2007spatial,sisbot2010synthesizing,trafton2005enabling,fong2006human}. While generating referring expressions, Magassouba et al. \cite{magassouba2019multimodal} addressed this problem by generating perspective-free expressions, i.e., expressions that don't tie to particular viewpoints. 

To generate unambiguous referring expressions, a robot needs to be explicit about from which perspective it refers to the objects, especially while using spatial relations between them. This can be achieved, for example, by clearly stating ``the object A that \textbf{I} see to the right of the object B'' or `the object A that \textbf{you} see to the left of the object B' to not to cause any misunderstanding during the interaction. 

While the robot is describing objects, taking the perspectives of users and evaluating its impacts have been remained unsolved until very recently. Lately, we have proposed a method for making the first attempt to address this problem. In our recent work \cite{dougan2020impact}, we observed a scene from different perspectives and generated an expression from the closest perspective of a user. Further, we evaluated the impact of perspective-taking with regard to different aspects of the interaction (e.g., task efficiency, perception of the task, and the robot). Through a user study, we showed that when the objects are spatially described from the users' perspectives, participants take less time to find the referred objects, find the correct objects more often and consider the task easier. 

Although our recent work demonstrates the significance of perspective-taking for effective collaboration, the method we proposed depended on the views of a scene from different perspectives. For a more general solution, i.e., taking different perspectives from a single view, 
3D or 6D pose information of the objects can be helpful. However, existing 3D or 6D off the shelf object pose predictors are still generally limited by predicting a few types of objects — mostly vehicles, pedestrians, trees. Therefore, they are not sufficient to use in real-world settings. Further, taking different perspectives in 2D is still an open topic in computer vision. Even though there are novel view synthesis methods \cite{liu2018geometry,sun2018multi,flynn2019deepview} that address this problem, existing solutions are still immature to use in real-world HRI.

Perspective-taking is also crucial when there are occluded objects from the views of users. When a robot observes a target object is occluded from the perspective of users, it should inform them that they need to change their viewpoint to see the target object and to achieve the task accurately. To address this problem, 3D cameras can be used, and occlusions can be estimated from the depth information of the objects. Further, if the problem is aimed to be resolved in 2D, studies on single-image depth estimation \cite{lee2018single,NIPS2016_6510,liu2015deep} can be employed to predict the depth of objects.



Although perspective-taking has been addressed while comprehending referring expressions and recently to generate these expressions, while the robot is describing objects, taking a user's perspectives from a single view and informing the user about occlusions is still open for further research. 

\subsection{Handling of Misinterpretations in an Autonomous Manner}

When a robot describes an object to its partner, it needs to cope with misinterpretations of its expressions and clarify them in an autonomous manner. In other words, when environments are highly ambiguous, users might misinterpret the expression of a robot (e.g., they might head towards the wrong objects) or ask for clarification. In these cases, the robot should be able to clarify its expressions in an autonomous manner to accurately describe the target object in dynamically changing real-world environments. 


To describe objects autonomously (i.e., generating referring expressions directly from scenes without requiring any prior knowledge about the environment, existing objects, or their configurations), recent studies in robotics have relied on deep learning \cite{dougan2019learning,magassouba2019multimodal,tanaka2019generating}. 
In our recent work \cite{dougan2019learning}, we have proposed a method to generate spatial referring expressions in a natural and unambiguous manner in real-world environments. 


In order to handle misinterpretations while a robot comprehends referring expressions, 
Shridhar and Hsu \cite{shridhar2018interactive,shridhar2020ingress} have proposed a method grounding by generation. They have used the generation part of their model for asking disambiguating questions during comprehension. Moreover, to handle misinterpretations while a robot refers to objects, Wallbridge et al. \cite{wallbridge2019generating} have proposed a dynamic method. In this method, there are three different dynamic description categories (i.e., negate, elaborate, positive) to provide user further clarification.

Even though these systems have made some advances to cope with misinterpretations,
they are either mainly focused on comprehension and a specific task (i.e., pick and place) \cite{shridhar2018interactive,shridhar2020ingress} or limited by the number of clarification categories \cite{wallbridge2019generating}. In order to handle misinterpretations while describing objects, existing autonomous referring expression generation methods need to be extended with detecting misinterpretations and providing more flexible clarifications. For this purpose, works on explainability (i.e., building more transparent and understandable models in their prediction-making process \cite{BarredoArrieta2020}),  visual question answering (i.e., generating an answer for a given image and a question \cite{antol2015vqa}), or visual dialog (i.e., generating an answer for a given image and a history of a dialog \cite{Das_2017_CVPR}) can be avenues worth exploring.




\section{Conclusion}
In this paper, we have focused on open challenges while generating referring expressions. We have suggested that utilizing contextual information and adapting expressions concerning context might contribute to more effective language-based interactions between robots and people. Further, we have stated that being explicit about from which perspective the expression is generated and taking the user perspective might be helpful for an unambiguous and efficient interaction. Finally, we have claimed that the handling of misinterpretations in an autonomous manner  
is necessary for successfully completing the task in dynamically changing environments.

\bibliographystyle{ACM-Reference-Format}
\bibliography{references}
\end{document}